\documentclass[journal]{IEEEtran}
\IEEEoverridecommandlockouts
\usepackage{cite}
\usepackage{amsmath,amssymb,amsfonts}
\usepackage{algorithmic}
\usepackage{graphicx}
\usepackage{textcomp}
\usepackage{xcolor}
\usepackage{cite}
\usepackage{amsmath,amssymb,amsfonts}
\usepackage{algorithm}
\usepackage{multirow}
\usepackage{siunitx}
\usepackage{algorithmic}
\usepackage{comment}
\usepackage{caption}
\usepackage{subcaption}
\usepackage{booktabs,
            makecell, % for second example
            tabularx}
\usepackage{textcomp}
\usepackage{mathtools}
\def\BibTeX{{\rm B\kern-.05em{\sc i\kern-.025em b}\kern-.08em
    T\kern-.1667em\lower.7ex\hbox{E}\kern-.125emX}}

\ifCLASSINFOpdf

\else

\fi

\begin{document}

\title{XMI-ICU: Explainable Machine Learning Model for Pseudo-Dynamic Prediction of Mortality in the ICU for Heart Attack Patients}

\author{Munib Mesinovic\textsuperscript{1}, Peter Watkinson\textsuperscript{2}, Tingting Zhu\textsuperscript{1}% <-this % stops a space
\thanks{\textsuperscript{1}Department of Engineering Science, University of Oxford, Oxford, UK}% <-this % stops a space
\thanks{\textsuperscript{2}Critical Care Research Group, Nuffield Department of Clinical Neurosciences, University of Oxford, Oxford, UK}% <-this % stops a space
}

\maketitle

\begin{abstract}
 \\

\noindent Heart attack remain one of the greatest contributors to mortality in the United States and globally. Patients admitted to the intensive care unit (ICU) with diagnosed heart attack (myocardial infarction or MI) are at higher risk of death. In this study, we use two retrospective cohorts extracted from the eICU and MIMIC-IV databases, to develop a novel pseudo-dynamic machine learning framework for mortality prediction in the ICU with interpretability and clinical risk analysis. The method provides accurate prediction for ICU patients up to 24 hours before the event and provide time-resolved interpretability results. The performance of the framework relying on extreme gradient boosting was evaluated on a held-out test set from eICU, and externally validated on the MIMIC-IV cohort using the most important features identified by time-resolved Shapley values achieving AUCs of 91.0 (balanced accuracy of 82.3) for 6-hour prediction of mortality respectively. We show that our framework successfully leverages time-series physiological measurements by translating them into stacked static prediction problems to be robustly predictive through time in the ICU stay and can offer clinical insight from time-resolved interpretability. \\
\end{abstract}

\begin{IEEEkeywords}
digital health, explainability, intensive care, machine learning, myocardial infarction, prediction, tabular deep learning
\end{IEEEkeywords}

\IEEEpeerreviewmaketitle

\section{MAIN}
\noindent Acute myocardial infarction (AMI) or heart attack is one of the greatest contributors to cardiovascular deaths in the world whose incidence remains critically high with approximately every 40 seconds someone in the United States suffering an episode \cite{tsao2022heart}. Cardiovascular diseases (CVDs) also represent a major cost burden globally with MI in the ICU being one of the most common CVD-related conditions in the critical care system \cite{degano2015twenty}. In 2015, there were more than 18 million CVD-related deaths with MI accounting for over 15\% of overall mortality and research showing that healthcare costs skyrocket with longer and more inefficient treatment in the ICU \cite{jayaraj2019epidemiology, roth2017global, soekhlal2013treatment}. A considerable amount of previous work was concerned with the classification and diagnosis of MI in the ICU with measurements using ECG signals or subtypes of MRI, but due to the acute nature of the condition and its urgent need for immediate therapy, these proposals have done little to proactively forecast the disease prior to occurrence, a task of high clinical relevance \cite{chen2022prediction}. Even the use of time-granular troponin assays, a biological marker for myocardial injury and thus infarction only helps with diagnosing the occurrence of an MI event faster but not with its prediction a priori \cite{than2019machine}. Therefore, prediction and timely treatment of MI as well as its risk factors in a high-risk population such as previous survivors is urgently needed and will not just help treat these vulnerable patients but will also help streamline the costs and burdens of the critical care system. \\

\noindent Patients who exhibit MI are usually referred to the ICU, however, they are 10\% more likely to suffer another episode in the days following and are at higher risk of death, especially the elderly \cite{nair2021characteristics}. Mortality prediction models can help design treatment plans and reduce costs and mortality rates but existing mortality prediction tools like the APACHE system deployed in US critical care centres have been criticised as too general and inaccurate for specific populations and diseases \cite{barrett2019building, venkataraman2018mortality}. One reason for machine learning's rise is its predictive performance compared to existing statistical and simple linear tools as well as its ability, in different cases and methods, of learning complex non-linear behaviours among variables \cite{mandair2020prediction}. When combined with interpretability methods, machine learning can be a useful tool for clinical guidance and decision-making. \\ 

\noindent Deep learning has been the core focus of the research community as it has shown incredible success in imaging and text problems, including in healthcare \cite{liu2019comparison}. The great advantage being that it does not require user-defined features and instead uses representational learning for tasks. Recent advances in tabular deep learning like TabNet and NODE have been a topic of lively conversation in the machine learning community with their interesting algorithmic and modular compositions but whether they can surpass classical machine learning models in different tasks is an ongoing debate \cite{joseph2021pytorch, gorishniy2021revisiting}. One of the drawbacks of such models is their opaqueness, lack of familiarity with tuning parameters, costs of training, and a dependency on a large amount of data being available. While deep learning models are the current standard in time-series EHR processing, we hope to show that by transforming the problem into connected and stacked static prediction problems, more reliable and low-cost models like extreme gradient boosted ensembles can be used instead and achieve superior performance to the deep learning alternatives. \\

Due to the recent nature of the proposed tabular deep learning models, research applying them to different healthcare challenges has been limited with only one recent paper looking at ICU mortality prediction with TabNet in COVID-19 patients specifically \cite{nazir2022interpretable}. Prior work, particularly research that used deep learning, has largely limited the populations studied in the ICU settings to either general admission populations, acute kidney injury, or sepsis patients \cite{shillan2019use, parreco2019comparing, moor2021early}. Other, more related work, has looked at survival and MI occurrence after ICU discharge, but again at a general patient population and without addressing the needs for prediction during ICU stay where the immediate risks of death for this population are considerable \cite{olsson2021prediction, law2002underlying}. \\

\noindent It is, therefore, both of interest and need to propose a machine learning framework that can reliably predict negative outcomes for heart attack patients in the ICU, test it independently, validate it externally, and provide useful interpretability of its predictions for clinicians. \\

\noindent Our contributions are as follows:

\begin{enumerate}
    \item We propose XMI-ICU, \textbf{X}GBoost for \textbf{M}yocardial \textbf{I}nfarction in the \textbf{ICU}, a novel gradient-boosted machine learning framework for the ICU that predicts mortality in MI patients, and that beats both existing prediction tools in active use as well as complex deep learning models recently proposed for tabular data
    \item We evaluate our model the high-risk ICU population of MI survivors using multiple ICU centres and externally validated in a separate ICU cohort
    \item We show the robustness of our model for varying time prediction in the ICU, including 6, 12, 18, and 24 hours in advance of the event of death
    \item We investigate the clinical risk factors most informative to the prediction and stratify them across time of stay using time-varying Shapley values in the ICU to show how different clinical attributes indicate risk at different times prior to the event
    \item We verify our model's clinical significance by a combination of interpretability methods like Shapley value analysis and clinical risk benefit and decision curves
\end{enumerate}

\section{RESULTS}
\subsection{\MakeLowercase{e}ICU}
\noindent Applying the framework proposed in Figure \ref{fig:Framework}, we compare our proposed XMI-ICU gradient-boosted model to standard supervised learning methods. For some of the methods like support vector machines (SVMs), we have standardised features. All features listed in the supplementary section on data processing were used for the eICU test results while only the most important (top 8) features identified by Shapley values analysis used for external validation on MIMIC-IV. APACHE IV score was not used as a feature in these models, albeit experiments doing so are included in the Supplementary Materials for those curious. The first set of results in Table \ref{tab:Result_MI} concerns prediction of mortality in MI patients with several hours prior to the event. It is clear that XMI-ICU maintains superior performance across all metrics for a priori prediction beating state-of-the-art tabular deep learning models. For AUROC and average precision, we evaluated the model at the default risk threshold in the results presented in the tables. Validation results include the mean and standard deviation of the 5 stratified folds. All XMI-ICU results have been checked for statistical significance (n=1000; p$<$.001). The results can also be seen visualised in Figure \ref{fig:MI_Results} which highlights the superior performance of XMI-ICU compared to alternative supervised learning models as measured by both AUROC and average precision. \\

\begin{table*}[t]
    \small
    \centering
    \caption{eICU validation (Val: Mean $\pm$ SD) and test prediction results for mortality prediction 6 hours in advance. Details on the metric computations can be found in the Supplementary Materials.}
    \label{tab:Result_MI}
    \begin{tabularx}{0.6\textwidth}{@{}l
                                    r
                                    r
                                    r
                                    r
                                 @{}}
    
        \toprule
         & Val AUC  & AUC  & Accuracy* & Average Precision\\
          \midrule
          XMI-ICU & \bf{91.8 $\pm$ 0.4} & \textbf{92.0} & \textbf{82.3} & \textbf{68.8}\\
          \midrule
          TabNet & 85.7 $\pm$ 2.1 & 84.1 & 77.0 & 60.7\\
          \midrule
          TabNet (pretrained) & - & 82.2 & 76.0 & 64.1\\
          \midrule
          NODE & 86.7 $\pm$ 0.7 & 85.4 & 67.6 & 62.3\\
          \midrule
          Logistic\\ Regression & 90.2 $\pm$ 0.4 & 89.6 & 73.5 & 61.5\\
          \midrule
          Random Forest & 91.0 $\pm$ 0.5 & 90.6 & 78.2 & 64.4\\
          \midrule
          SVM & 90.2 $\pm$ 0.8 & 89.3 & 77.0 & 58.1\\
          \midrule
          SVM (linear) & 87.4 $\pm$ 0.7 & 87.7 & 78.8 & 63.8\\
          \midrule
          LDA & 79.4 $\pm$ 2.0 & 78.7 & 51.0 & 29.3\\
            \bottomrule
    \end{tabularx}
\end{table*}

    \begin{figure*}
        \centering
        \begin{subfigure}[b]{0.475\textwidth}   
            \centering 
            \includegraphics[width=\textwidth]{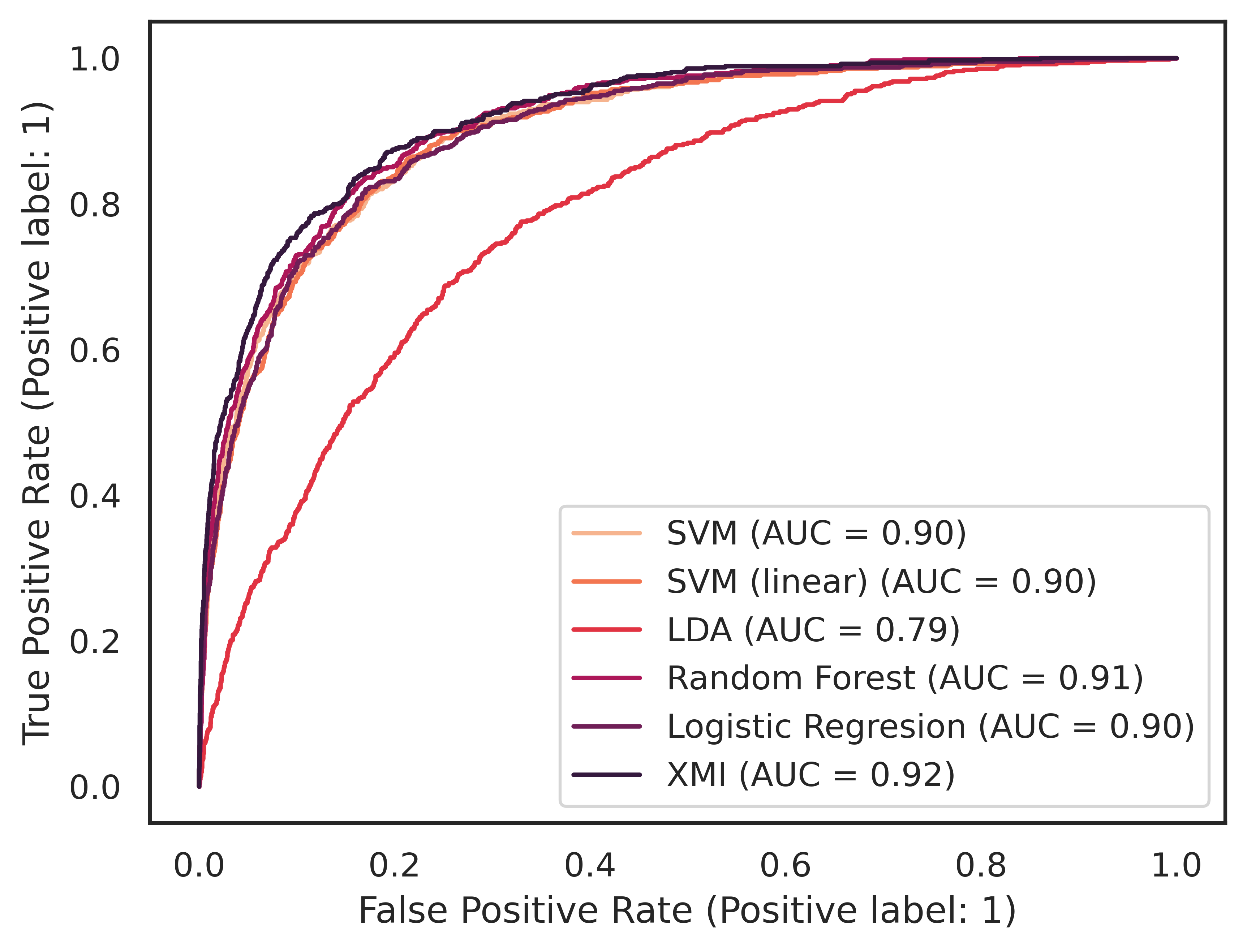}
            \caption[]%
            {{\small AUROC performance of models for mortality prediction}}    
            \label{fig:mean and std of net34}
        \end{subfigure}
        \hfill
        \begin{subfigure}[b]{0.475\textwidth}   
            \centering 
            \includegraphics[width=\textwidth]{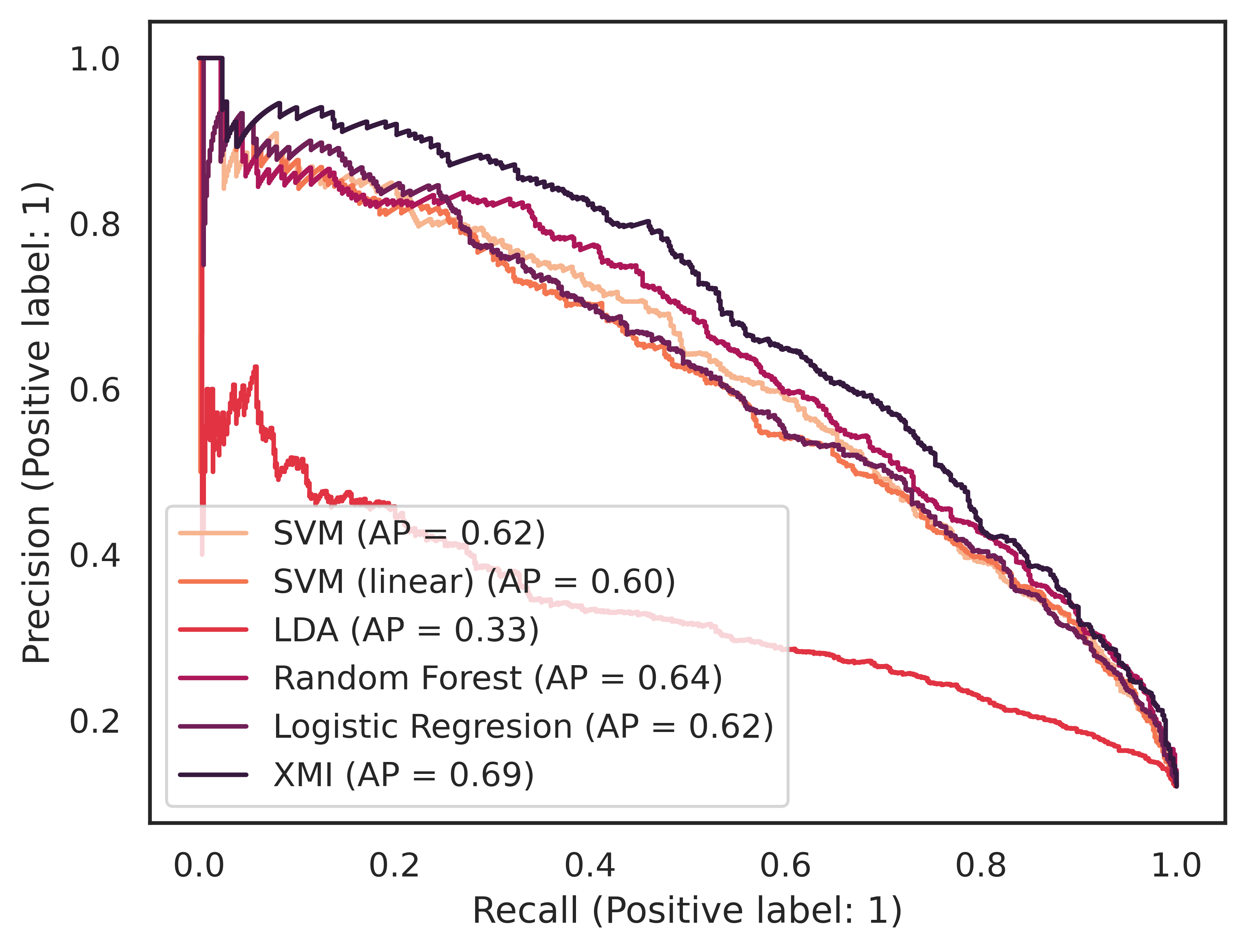}
            \caption[]%
            {{\small AP of models for mortality prediction}}
            \label{fig:mean and std of net44}
        \end{subfigure}
        \caption
        {\small Evaluation performance of XMI-ICU to predict mortality 6 hours in advance compared to other models for different metrics on eICU held-out test. AUROC and average precision results are measured across prediction thresholds and they support the claim that XMI-ICU is a successful prediction model beyond default threshold values.} 
        \label{fig:MI_Results}
    \end{figure*}

\noindent After the XMI-ICU model was evaluated at 6 hour prediction prior to death, we extend to a more dynamic prediction evaluation by adapting the framework to arbitrarily predict the events of death at any time prior, and the framework will automatically extract, preprocess, standardise existing measurements, optimise respective hyperparameters, and deploy the XGBoost model for test prediction. The results for XMI-ICU evaluated at 6, 12, 18, and 24 hour prediction for mortality in held-out test set of eICU can be seen in Table \ref{tab:Result_MI_Time} and they continue to show reliable predictive performance across the different time windows. The table also includes the results for APACHE-IV as a matter of comparison for 24 hour prediction of mortality. \\

\begin{table*}[t]
    \centering
    \caption{eICU test with all features, eICU test only using top 8 features, and MIMIC-IV external validation (Val: Mean $\pm$ SD) prediction results for mortality prediction stratified with time for XMI-ICU. External validation uses all eICU data as train set and MIMIC-IV data as test set with only the top 8 features included as identified by Shapley value analysis. Accuracy stands for balanced accuracy, details on the metric computations can be found in the Supplementary Materials.}
    \label{tab:Result_MI_Time}
    \setlength{\tabcolsep}{10pt}
    \begin{tabularx}{0.7\textwidth}{l
                                    r
                                    r
                                    r
                                    r
                                 }
    
        \toprule
         & Val AUC  & AUC  & Accuracy* & Average Precision\\
         \midrule
         \textbf{eICU Mortality} &&&& \\
          \midrule
          6 hours & 91.8 $\pm$ 0.4 & 92.0  & 82.3 & 68.8\\
          \midrule
          12 hours & 90.5 $\pm$ 0.7 & 89.9 & 81.9 & 65.8\\
          \midrule
          18 hours & 89.1 $\pm$ 1.0 & 89.8 & 81.2 & 65.5\\
          \midrule
          24 hours & 87.7 $\pm$ 1.0 & 88.2 & 80.4 & 63.0\\
          \midrule
           \midrule
          APACHE IV & - & 69.9 & 69.3 & 31.5\\
          \midrule
          \midrule
          \textbf{Top-8 eICU Mortality} &&&& \\
          \midrule
          6 hours & 86.7 $\pm$ 1.1 & 86.2 & 80.0 & 74.7\\
          \midrule
          12 hours & 85.2 $\pm$ 1.2 & 83.3 & 77.0 & 69.7\\
          \midrule
          18 hours & 83.4 $\pm$ 1.3 & 83.1 & 76.5 & 65.8\\
          \midrule
          24 hours & 81.9 $\pm$ 1.4 & 81.2 & 75.2 & 59.2\\
          \midrule
          \textbf{External MIMIC-IV Mortality} &&&& \\
          \midrule
          6 hours & - & 80.0 & 77.7 & 73.8\\
          \midrule
          12 hours & - & 77.7 & 75.9 & 69.9\\
          \midrule
          18 hours & - & 76.6 & 75.1 & 67.8\\
          \midrule
          24 hours & - & 75.1 & 74.9 & 66.5\\
            \bottomrule
    \end{tabularx}
\end{table*}

\begin{figure*}
        \centering
        \begin{subfigure}[b]{0.475\textwidth}  
            \centering 
            \includegraphics[width=\textwidth]{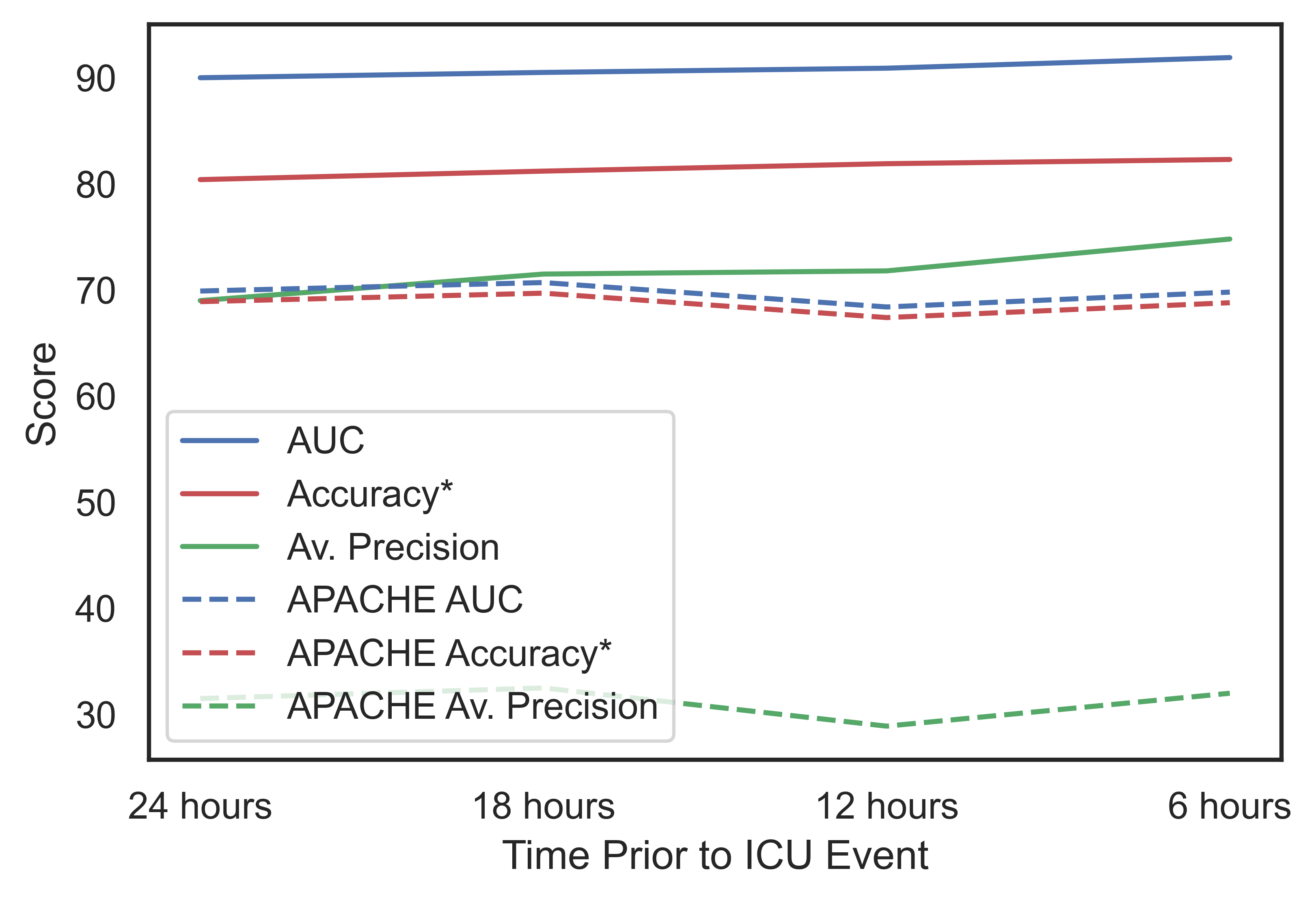}
            \caption[]%
            {{\small XMI-ICU performance across time for mortality predictions on eICU held-out test set and APACHE performance (dotted)}}
            \label{fig:Time_XMI-ICU2}
        \end{subfigure}
        \begin{subfigure}[b]{0.475\textwidth}  
            \centering 
            \includegraphics[width=\textwidth]{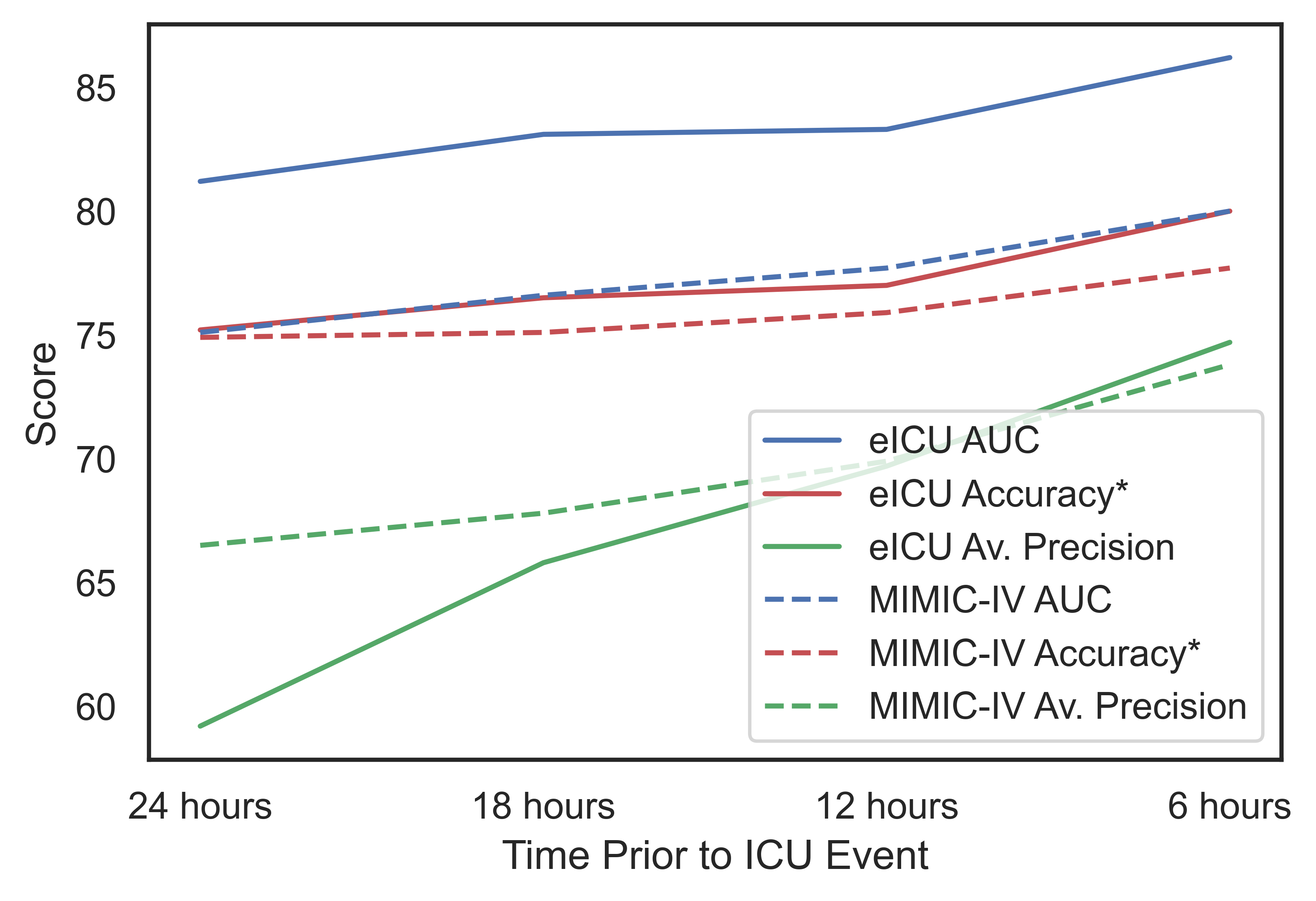}
            \caption[]%
            {{\small Performance across time for mortality prediction on eICU test set and external MIMIC-IV set (dotted) with only top 8 features}}    
            \label{fig:Time_XMI-ICU}
        \end{subfigure}
        
        \caption
        {\small Robustness and reliability of XMI-ICU prediction performance over time in the ICU for mortality prediction (left) using all features available in eICU and as measured by a variety of metrics. The right figure contains results from eICU held-out test set and MIMIC-IV external cohort with only the top 8 features identified by Shapley value analysis.} 
        \label{fig:MI_Results}
    \end{figure*}

\noindent A plot showing stability of predictive performance across different metrics for XMI-ICU as a function of time in the ICU prior to death can be seen in Figure \ref{fig:Time_XMI-ICU2}. Figure \ref{fig:Time_XMI-ICU2} indicates stable performance for mortality prediction and its superiority compared to APACHE-IV (included for times beyond just 24 hours). The right figure shows the generalisation ability of the model to perform on a completely external test set which is MIMIC-IV using only top 8 features from eICU for training. Evaluating a time-prediction model like XMI-ICU also requires showing coherent prediction across time and not just consistency of prediction accuracy and robustness. In the next set of results, we show XMI-ICU with low misclassification error across time for the same patient sample. A patient is deemed misclassified if they are predicted incorrectly at time x in advance when they have been previously predicted correctly at times $>$x. For example, a patient might be predicted to die at the 24 and 18 hour prediction windows correctly but at 12 hours in advance they are predicted (incorrectly) to survive. These instabilities in prediction across time need to be measured if the model is to sustain reliable performance throughout the ICU stay. We define three patient subcohorts as illustrated in the top of Table \ref{tab:Result_Death_Time_Error} where each indicates the group of patients correctly predicted at all previous time windows except one. The bottom of Table \ref{tab:Result_Death_Time_Error} presents these results for both death and heart attack prediction indicating the low levels of misclassification most likely indicate sensitivity to noise rather than predictive weakness. \\

% \begin{table*}[t]
% %    \setlength{\tabcolsep}{7pt}
%     \small
% %    \renewcommand*{\arraystretch}{0.7}
%     \centering
%     \caption{Defined patient cohorts for evaluating XMI-ICU predictive robustness across time windows. Each patient cohort corresponds to a grouping of patients who have been wrongly predicted at time x after being correctly predicted at all times before.}
%     \begin{tabularx}{0.5\textwidth}{l
%                                     c
%                                     c
%                                     c
%                                     c
%                                  }
%          \toprule
%          Patient Cohort & 24 hours  &  18 hours  &  12 hours & 6 hours\\
%             \midrule
%           \textbf{${P_{1}}$} & \checkmark & \checkmark &  \checkmark & X\\
%           \midrule
%           \textbf{${P_{2}}$} & \checkmark & \checkmark & X & \\
%           \midrule
%           \textbf{${P_{3}}$} & \checkmark & X &  & \\
%             \bottomrule
%     \end{tabularx}
%     \label{tab:patient_cohorts}
% \end{table*}

\begin{table*}[t]
    \centering
    \caption{TOP: Defined patient cohorts for evaluating XMI-ICU predictive robustness across time windows. Each patient cohort corresponds to a grouping of patients who have been wrongly predicted at time x after being correctly predicted at all times before. BOTTOM: Misclassification rate (in percentage) is defined as number of wrong classifications divided by total patient sample present in cohorts for 6, 12, 18, and 24 hours prediction windows. A misclassification example is one where a patient is wrongly predicted in a time prediction window after being correctly predicted at previous windows.}
    \label{tab:Result_Death_Time_Error}
    \setlength{\tabcolsep}{15pt}
    \begin{tabularx}{0.6\textwidth}{l
                                    c
                                    c
                                    c
                                    c
                                 }
             \toprule
         Patient Cohort & 24 hours  &  18 hours  &  12 hours & 6 hours\\
            \midrule
          \textbf{${P_{1}}$} & \checkmark & \checkmark &  \checkmark & X\\
          \midrule
          \textbf{${P_{2}}$} & \checkmark & \checkmark & X & \\
          \midrule
          \textbf{${P_{3}}$} & \checkmark & X &  & \\
        \toprule
        \toprule
         &  ${P_{3}}$ &  ${P_{2}}$ &  ${P_{1}}$ &\\
          \midrule
          \textbf{eICU} & & & &\\
          \midrule
          Mortality & 7.9 & 8.2 & 5.5 &\\
          \midrule
          \textbf{MIMIC-IV} & & & &\\
          \midrule
          Mortality & 6.4 & 6.3 & 4.7 &\\
            \bottomrule
    \end{tabularx}
\end{table*}

\noindent To understand how XMI-ICU is making these predictions and obtain further analysis for clinical significance testing, we applied Shapley value analysis on the held-out test set and observe relative feature importance. We did so across all time window prediction problems with the 6 hour example found in the Supplementary. We also subjected our interpretability results to random perturbation tests by adding a Gaussian distributed feature to the feature set to evaluate the susceptibility of change in the top variables identified and we observe no significant changes by introduction of noise variables. An example of this test is also included in the Supplementary. \\

\begin{figure*}[t]

\centering

\includegraphics[width=0.67\linewidth]{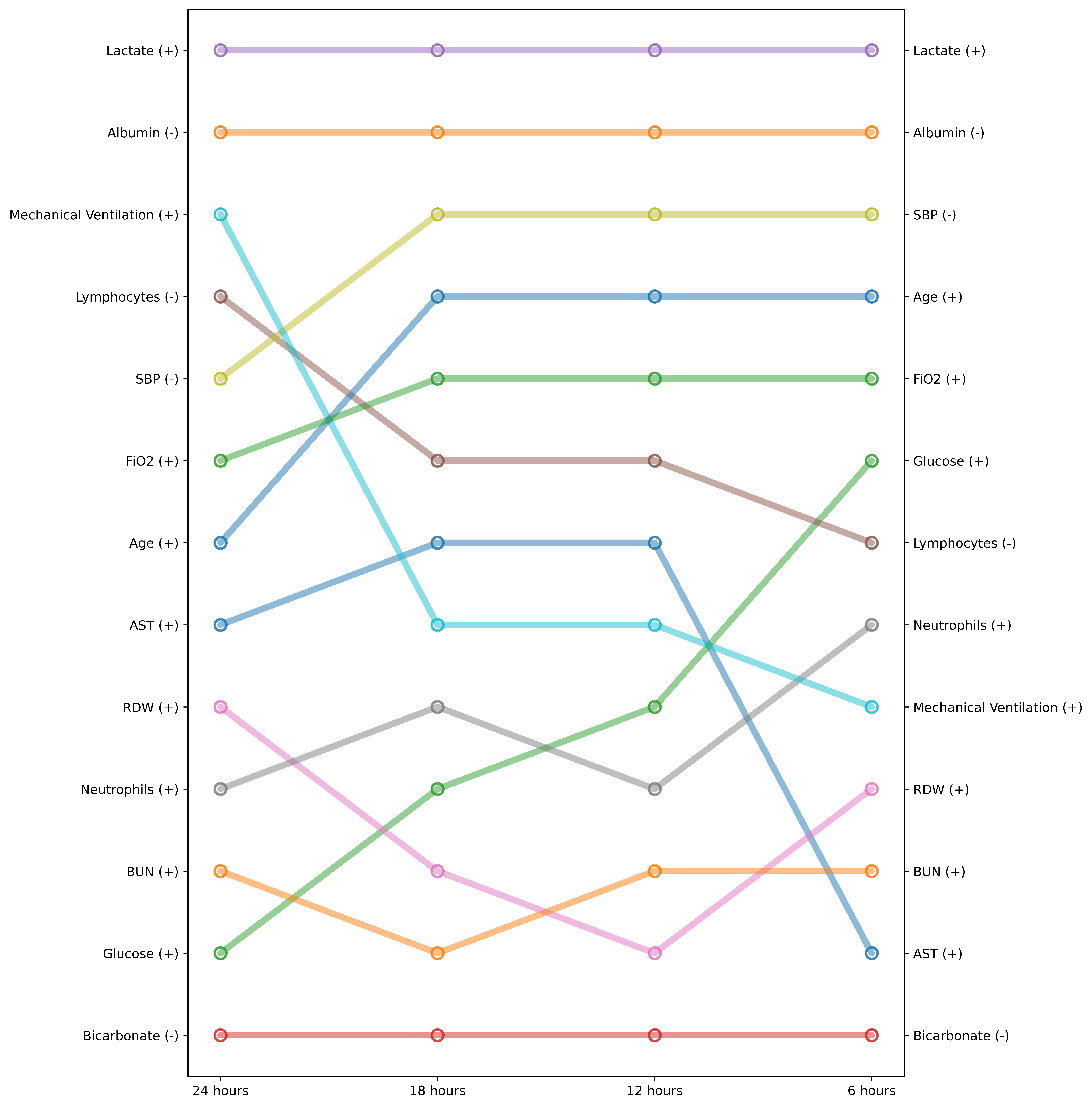}

\caption{Ranking of most important features as identified by their relative SHAP values for XMI-ICU prediction of mortality varied across time during ICU stay prior to event. For the time windows in the {6, 12, 18, 24} hour intervals, the top 13 features in each of the windows are presented as extracted from eICU thereby showcasing how the most important features for correct prediction of mortality changes through time or closer to the prediction event.}
\label{fig:Time_Shap}
\end{figure*}

\noindent We further stratify Shapley values as a function of time in the ICU for mortality prediction. A feature ranking at each time-point corresponds to the relative importance of that feature at that point in time in the ICU stay prior to the event in question. The time-graphs can be seen in Figure \ref{fig:Time_Shap}. These values were extracted for each of the time windows, in effect converting a static interpretability method to a dynamic explainability framework that shows how at different times closer to the event (death or heart attack) different values of features and their importance changes and how that is used by the model to learn underlying patters for disease outcome prediction. \\

\subsection{External Validation: MIMIC-IV}
\noindent We evaluated XMI-ICU on the separate and independent MIMIC-IV dataset for mortality prediction in MI patients. The features identified as most important by Shapley values analysis were used to create a new training set of the entirety of eICU and test on the entirety of MIMIC-IV cohorts only using the top 8 features whose statistical distributions for the different sets are included in Table \ref{tab:Characteristics1}. XMI-ICU maintains high predictive performance across metrics when tested on this external dataset as can be seen in Table \ref{tab:Result_MI_Time} without any training or tuning on it using only the top 8 features identified by Shapley value analysis from eICU test set. The results immediately above correspond to held-out test set performance for eICU using those same 8 features. \\

\noindent A plot showing predictive performance across different metrics for XMI-ICU evaluated on the MIMIC-IV cohort can be seen in the bottom Figure \ref{fig:Time_XMI-ICU}. We also evaluate XMI-ICU for 6-hour prediction across subpopulations due to our multi-centre diverse dataset across sex and ethnicity demographics as a fair robustness check. The results can be seen in Table \ref{tab:Subpopulation} showing stable performance for XMI-ICU across different subcohorts for both eICU and MIMIC-IV held-out test sets. \\

\begin{table*}[t]
    \centering
    \caption{AUROC test results for XMI-ICU evaluated on subpopulations for 6 hour prediction.}
    \label{tab:Result_demo}
    \setlength{\tabcolsep}{6pt}
    \begin{tabularx}{0.47\textwidth}{l
                                    r
                                    r
                                 }
    
        \toprule
           & Mortality & Mortality (external MIMIC-IV) \\
          \midrule
          Men  & 90.2 & 81.9\\
          \midrule
          Women & 92.7 & 77.5 \\
          \midrule
          Caucasian & 91.7 & 81.8 \\
          \midrule
          Black/Hispanic & 92.3 & 75.6\\
            \bottomrule
    \end{tabularx}
    \label{tab:Subpopulation}
\end{table*}

\subsection{Clinical Risk Benefit Analysis}
\begin{figure*}
        \centering
            \includegraphics[width=\textwidth]{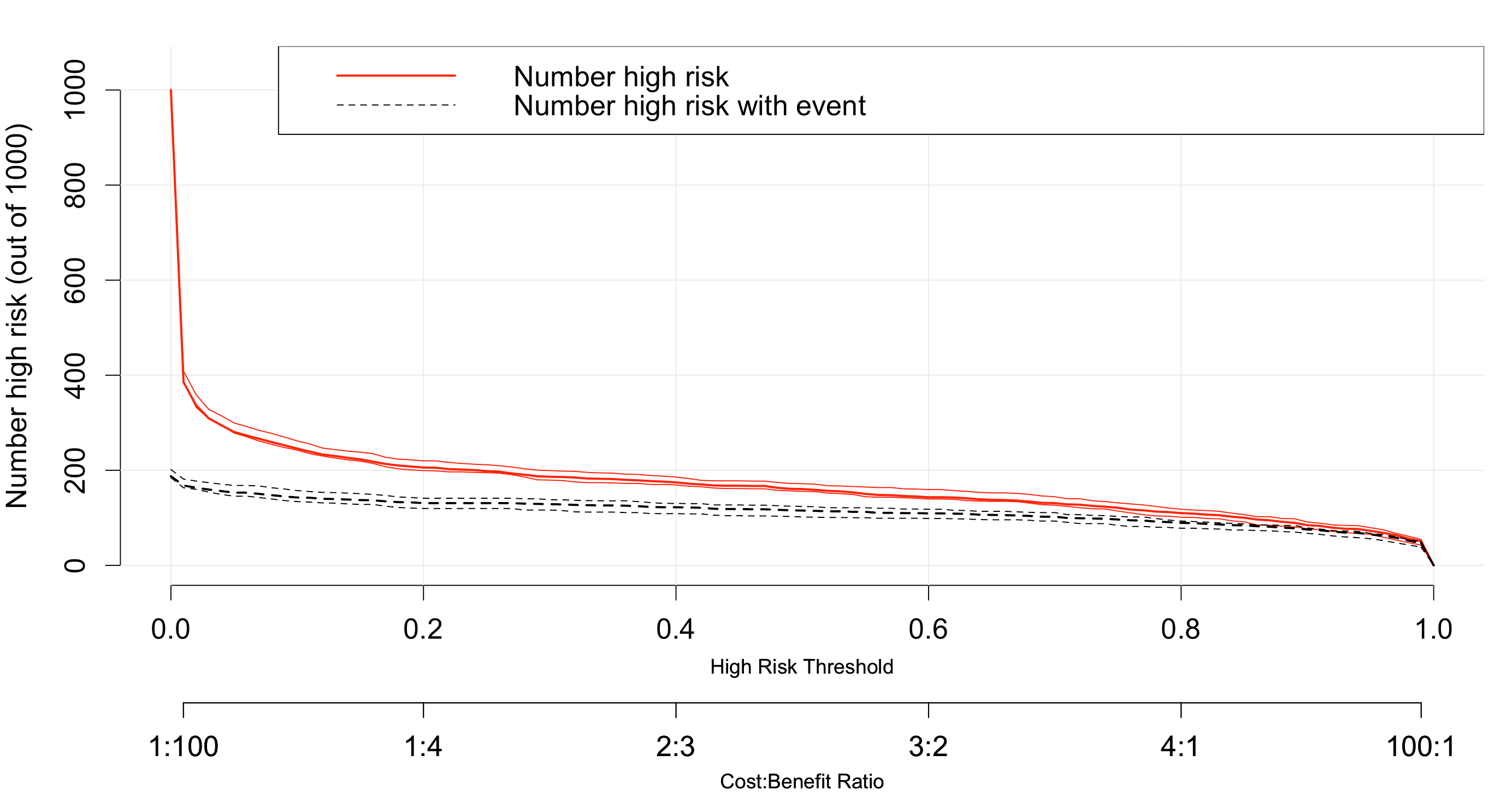}
        \caption
        {\small Clinical decision-making evaluation performance of XMI-ICU for mortality prediction using only the top 8 features on the entire eICU test set. Here we include the clinical impact curve measuring the risk predicted by XMI-ICU across different risk groups relative to the actual risk. For each risk threshold, we see the propensity of our prediction model to over- or underestimate risk of that event.} 
        \label{fig:CIC}
    \end{figure*}

\begin{figure*}
            \centering 
            \includegraphics[width=\textwidth]{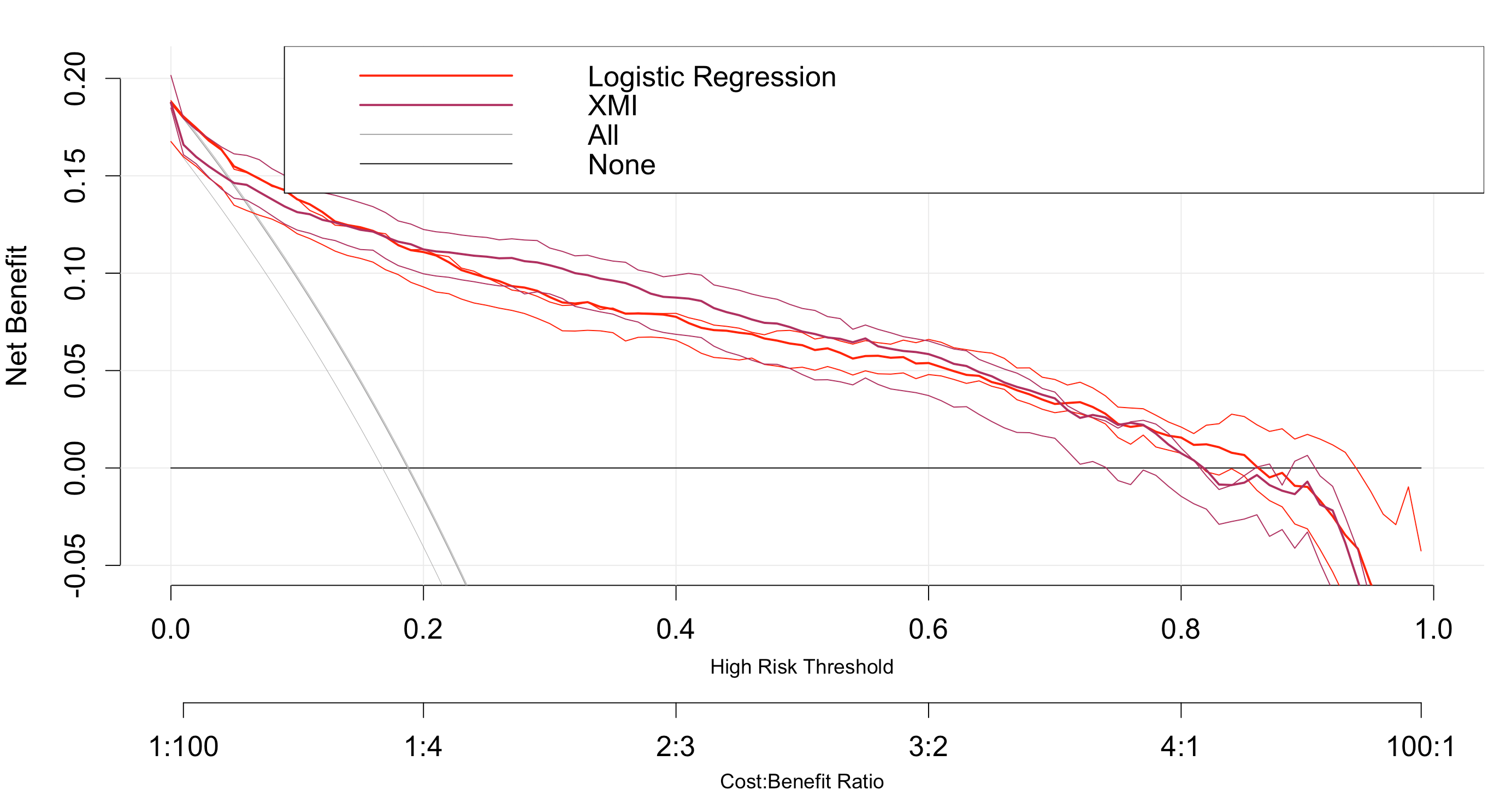}
        \caption
        {\small Clinical decision-making evaluation performance of XMI-ICU for mortality prediction using only the top 8 features on the entire eICU test set. Here we include decision curves comparing the net benefit of XMI-ICU to logistic regression (analog to APACHE IV) models across risk groups as defined by the risk thresholds. In the decision curves, the "All" tag corresponds with the net benefit behaviour of having all patients predicted positive and "None" with having no patients predicted positive.} 
        \label{fig:DC}
    \end{figure*}

\noindent To communicate the clinical significance of the XMI-ICU model results to clinicians, we evaluated our model with clinical impact curves (Figure \ref{fig:CIC}) and decision curve estimates (Figure \ref{fig:DC}) for robust risk evaluation. A 90 percent confidence interval was derived with 50 bootstrap iterations on the test set. As the clinical impact curves for mortality show, XMI-ICU consistently identifies patients at risk across different risk thresholds showing robustness to false negatives. For those at highest risk ($>$75\%), XMI-ICU has very low tendencies for false positives or "over-risking" in its predictions, learning to focus on those most at risk with higher specificity and sensitivity. The decision curves indicate XMI-ICU's approximated net benefit outperforming logistic regression (underlying model used in APACHE) using only top features identified from Shapley values analysis. \\

\noindent To assist clinicians more readily in their decision making process, the top features of our XMI-ICU model were used to construct a nomogram which is included in the Supplementary Materials and shows a simple representation of what could be an automatic calculator for 24-hour risk calculation in the ICU. \\

\section{DISCUSSION}
\noindent Our proposed XMI-ICU model shows superior predictive performance for mortality prediction. Since our prediction time for mortality is at least 6 and up to 24 hours before the events but works for any arbitrary time that leaves clinicians with flexible extra time to prioritise high risk patients and administer preventative measures. Interestingly, both TabNet and NODE do not achieve high performance when compared to XMI-ICU. These results directly contribute to the ongoing debate on the comparisons of tabular deep learning with classical methods which have shown mixed results over the last year in published research \cite{fayaz2022deep, shwartz2022tabular}. We provide the first set of comparisons for these models in healthcare data known to be noisy, sparse, and presenting unique challenges. From our results it is clear that both complex and costly training and optimisation and longer deployment times combined with lackluster performance makes these tabular deep learning models currently incomparable to gradient-boosted methods like XMI-ICU. \\

\noindent XMI-ICU also beats the existing prediction tool in use across ICUs in the United States, APACHE IV, by 18.3\% in test AUROC and 11.1\% in test accuracy at 24-hour prediction. It requires miliseconds to be deployed once trained which also only takes a couple of seconds allowing for rapid response times in the ICU. Additionally, as Figure \ref{fig:Time_XMI-ICU2} shows, XMI-ICU maintains stable performance across all metrics during the 24 hours of ICU stay prior to death for MI patients. The model also successfully performs mortality prediction across different prediction time-windows in an external patient cohort obtained from MIMIC-IV using only the 8 most important features identified by Shapley values analysis on eICU as seen in Figure \ref{fig:Time_XMI-ICU}. The drop in predictive performance compared to using MIMIC-IV as part of the training set with all features included is expected as we now use only 8 features without any training on the MIMIC-IV dataset itself. Despite this challenge, XMI-ICU maintains relatively high external predictive performance. \\

\noindent XMI-ICU combined with interpretability provides clinical risk factor importance which can aid physicians in both relying on the model but also investigating what aspects of the physiological measurements are more informative at what time during the ICU stay. For mortality prediction, we see that the closer a patient gets to highest risk of death close to the event, whether they are mechanically ventilated drops in importance compared to blood measurements like higher lactate, lower albumin, and systolic blood pressure. Hyperlactatemia has been found to be highly associated with in-hospital mortality in a relatively small and isolated heterogeneous ICU population \cite{van2013cumulative}. Our findings on a much larger multi-centre patient cohort provide predictive evidence for this case. Previous research has established that lower serum albumin levels are good predictors of higher risk of death in ICU patients with sepsis and COVID-19 while our work seems to suggest a similar predictive pattern for heart attack patients as well \cite{kendall2019serum, aziz2020association}. While this is currently a matter of debate we are glad to contribute to in medical sciences, lower albumin levels may be a marker of persistent injury to arteries and progression of atherosclerosis and thrombosis. The more time passes from low albumin levels the higher the risk of further acute injury in the myocardium which is what would make it useful for tracking risk of MI as our results suggest \cite{djousse2002serum, oduncu2013prognostic}. \\

\noindent Prior work has shed light on the hypothesis that hypotension as measured by the lowering of systolic blood pressure can be an indicator of higher risks of death in ICU patients, specifically those with acute kidney injury \cite{li2013methods}. Some have suggested that myocardial injury is also more likely in cases of lower SBP values but here we provide early indication of the high prediction value of lower SBP levels for heart attack in the ICU \cite{maheshwari2018relationship}. The sudden rise in high glucose levels and variability (as captured by our standard deviation measure for blood glucose) being strong predictors of mortality have been confirmed with several retrospective cohort studies in the ICU \cite{hermanides2010glucose, gunst2019glucose}. Work on prioritising those patients with such measurements and controlling for blood glucose and albumin can more easily be extended to preventive care for MI patients as well. \\

\noindent Comparing the framework to existing deep learning time-series models that tend to be costly and complex, our system with its simple embedded gradient boosted model sensitive to class imbalance and with dynamic feature extraction maintains prediction fidelity at varying time points while being faster, more interpretable, and less environmentally and financially costly to train and deploy. The XMI-ICU dynamic framework also offers an alternative to the rush in clinical machine learning in applying costly and less intepretable transformer and time-series models to these types of problems while still providing a dynamic prediction framework. \\

\noindent In conclusion, we developed a highly predictive machine learning framework that trains on time-series ICU ward data without requiring complex deep learning models. Instead, it relies on dynamic feature extraction and takes advantage of the predictive power of static models like XGBoost which outperformed other models including state-of-the-art tabular deep learning. The framework offers time-resolved interpretability that allows tracking changes in vital sign and blood measurement importance across the ICU stay for heart attack patients whose conclusions seek to provide medical insight. The framework could be integrated into ICU systems to predict negative outcomes in heart attack patients with real-time patient measurements. \\

\section{METHODS}
\subsection{Study design and population}
\noindent The data used in this study is the eICU Collaborative Research Database is a public database available upon request and fulfillment of ethical training \cite{pollard2018eicu}. The eICU database was processed using postgreSQL and the \textit{pandas} package. eICU  is a multi-center ICU database with over 200,859 patient unit encounters for 139,367 unique patients admitted between 2014 and 2015 to one of 335 ICUs at 208 hospitals located throughout the United States \cite{pollard2018eicu}. The database is de-identified and includes vital sign measurements, demographic data, and diagnosis information. For a full list of features used in our study please consult the relevant tables in the Supplementary Materials. \\

\begin{figure*}[t]

\centering
\subfloat[Cohort selection for eICU database]{
\centering
\includegraphics[width=0.4\linewidth]{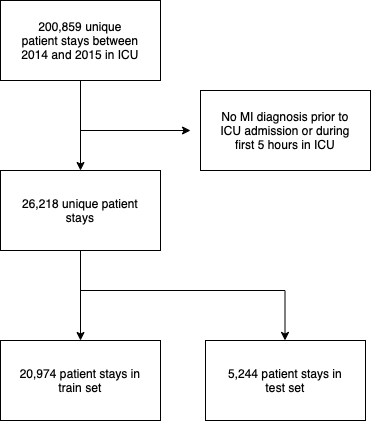}
}
%no space
\hfill
\subfloat[Cohort selection for MIMIC-IV database]{
\centering

\includegraphics[width=0.39\linewidth]{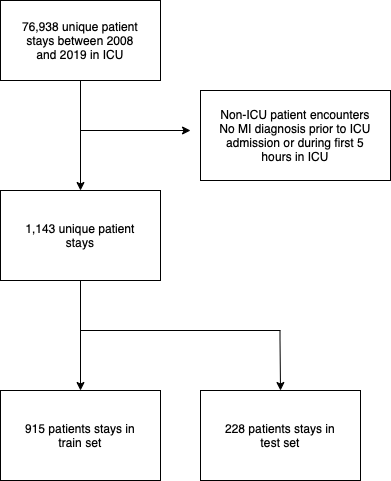}
}
\caption{MI patient cohort selection. The exclusion criteria were listed here as they were implemented in PostgreSQL and Pandas. The final exclusion criteria is to extract the relevant subcohort at the end which is MI admitted patients to the ICU.}
\label{fig:flowchart}
\end{figure*}

\noindent We based this study on the data preprocessing workflow used in \cite{rocheteau2021temporal}, but adapted it to our problem accordingly. Our inclusion criteria were patients of age$>$18 and $<$89 years with an ICU length of stay of at least 5 hours to remove transient patients. We also include those with at least one recorded observation and excluded those without any laboratory measurements. Patients on respiratory support had a separate set of measurements which we included with a mechanical ventilation tag feature for this patient subgroup. We included variables present in at least 12.5\% of patient stays, or 25\% for lab variables due to their relative sparsity. We then removed those patients without any diagnosis information after 5 hours of stay because they might be inactive ICU patients logged for longer than was the case. A similar approach was taken by \cite{sheikhalishahi2020benchmarking}. Our final subcohort consisted of 26,218 patients. We extracted diagnoses entered less than 5 hours after entering the ICU and diagnoses prior to admission as starting diagnosis or first diagnosis. A flowchart of the patients cohort selection can be seen in Figure \ref{fig:flowchart}. The minimum length of stay changes depending on what model prediction time one enters into the framework. For example, evaluating the model for 6-hour or 24-hour prediction time means that the minimum length of stay for those patients would have to be 6 or 24 hours for there to be existent measurements for analysis. \\

\noindent The eICU database as well as many of the ICUs in the United States use the APACHE IV system for mortality risk prediction. The Acute Physiology, Age, and Chronic Health Evaluation (APACHE) IV system is a tool used to risk-adjust ICU patients which provides estimates of the probability that a patient dies given data from the first 24 hours \cite{zimmerman2006acute}. The APACHE IV score is the result of a multivariate logistic regression which uses demographic, laboratory measurement, and diagnosis data to make a mortality risk assessment. It is the standard benchmark for mortality prediction tools in the ICU. Here we will use it both as a feature and as a benchmark as it was highly important for us to evaluate the score's feature importance downstream in our models. We will provide XMI-ICU prediction performance for 24 hours which is the most directly comparable to APACHE-IV. APACHE-IV is only present in the eICU dataset. \\

\noindent We define our myocardial infarction tag as a collection of the diagnosis strings and respective ICD-10 codes described in the Supplementary Material. Once we have the defined outcome as MI, we are left with 26,218 samples that have had a diagnosis of MI prior to admission to the ICU with 3,139 (12.0\%) having died during their stay. \\

We externally validated our model on the most recent release of Medical Information Mart for Intensive Care (MIMIC-IV v. 2.0, July 2022) which includes discharge information for over 15,000 additional ICU patients compared to the previous release \cite{johnson2020mimic}. Similar to eICU, MIMIC-IV is a de-identified and real world intensive care database using data from the Beth Israel Deaconess Medical Center for the years 2008 - 2019. We use similar cohort selection criteria as illustrated in Figure \ref{fig:flowchart} and label definition as in eICU resulting in 1,143 unique patient ICU stays with confirmed MI out of 76,938. 131, or 12.0\%, have died during their stay. Regarding missingness of variables, we mimic the steps taken for eICU processing. \\

The data processing of time-series and static variables was completed in Python. Patient cohort characteristics can be seen in Table \ref{tab:Characteristics1}. \\

\begin{table*}[t]
    \small
    \centering
    \caption{Summary of demographics and variables used for external validation across training and testing datasets. MIMIC-IV has been used separately as an external validation source with the summary statistics for the entire dataset being a compound average of its train and test set statistics listed here individually.}
    \begin{tabularx}{0.8\textwidth}{l
                                    r
                                    r
                                    r
                                    r
                                    r
                                    r
                                 }
         \toprule
        & \multicolumn{2}{c}{eICU (N = 26,218)} && \multicolumn{2}{c}{MIMIC-IV (N = 1,143)}  \\
        \cmidrule(r){2-3} \cmidrule(l){5-6}
         \bf{Attributes}  &  \multicolumn{1}{c}{\bf{Train (N = 20,974)}} & \bf{Test (N = 5,244)} &&  \multicolumn{1}{c}{\bf{\bf{Train (N = 915)}}} & \bf{Test (N = 228)}\\
            \midrule
          Age (mean $\pm$ SD) & 66.8 ($\pm$ 12.7) & 67.2 ($\pm$ 12.4) && 68.1 ($\pm$ 13.2) & 68.0 ($\pm$ 13.1)\\
          \addlinespace[0.05cm]
          Sex (male) & 13,369 (63.7\%) & 3,385 (64.5\%) && 585 (51.9\%) & 156 (55.4\%)\\
          \addlinespace[0.05cm]
          LoS (days) & 4.1 ($\pm$ 2.7) & 4.0 ($\pm$ 2.3) && 3.7 ($\pm$ 2.9) & 3.2 ($\pm$ 3.1)\\
          \addlinespace[0.05cm]
          Lactate & 2.9 ($\pm$ 2.8) & 2.5 ($\pm$ 2.3) && 2.0 ($\pm$ 1.5) & 1.9 ($\pm$ 1.5)\\
          \addlinespace[0.05cm]
          SBP & 120.2 ($\pm$ 17.9) & 120.0 ($\pm$ 16.3) && 126.3 ($\pm$ 18.8) & 124.5 ($\pm$ 13.1)\\
          \addlinespace[0.05cm]
          Glucose & 150.4 ($\pm$ 61.7) & 147.3 ($\pm$ 56.7) && 136.5 ($\pm$ 49.3) & 133.7 ($\pm$ 45.1)\\
          \addlinespace[0.05cm]
          WBC & 15.5 ($\pm$ 10.5) & 15.1 ($\pm$ 9.3) && 10.6 ($\pm$ 7.4) & 10.5 ($\pm$ 7.4)\\
          \addlinespace[0.05cm]
          RDW & 15.1 ($\pm$ 2.2) & 15.0 ($\pm$ 2.0) && 14.4 ($\pm$ 2.1) & 14.2 ($\pm$ 2.0)\\
          \addlinespace[0.05cm]
          Urea Nitrogen & 27.4 ($\pm$ 19.5) & 22.8 ($\pm$ 13.4) && 22.8 ($\pm$ 17.0) & 21.3 ($\pm$ 14.6)\\
          \addlinespace[0.05cm]
          Bicarbonate & 24.7 ($\pm$ 4.2) & 24.8 ($\pm$ 4.4) && 23.3 ($\pm$ 3.1) & 23.0 ($\pm$ 3.0)\\
          \addlinespace[0.05cm]
          Mortality (dead) & 2,511 (12.0\%) & 628 (12.0\%) && 105 (11.5\%) & 26 (11.3\%)\\
            \bottomrule
    \end{tabularx}
    \label{tab:Characteristics1}
\end{table*}

\subsection{Machine learning methods}
\noindent Following extraction of patients, we split the dataset into training and testing (20\%) with the test set being used as hold-out for reporting only the final results. The training set was used for hyperparameter tuning of different machine learning and deep learning models using either Bayesian optimisation (to help reduce the overall computational costs of the framework) or the traditional grid search with 5-fold stratified cross validation. The validation scores in the results section represent the results of this cross validation. The next step in the framework is to pad the missing measurements for the time-windows using imputation with Multivariate Imputation by Chained Equation (MICE) and for feature standardisation or normalisation where necessary to avoid any data leakage either inside the validation folds or, at the end, the held-out test set with the parameters extracted only on the training set or the training folds respectively \cite{zhang2016missing}. Instead of using resampling techniques like SMOTE which can incur bias, we use inverse class-weighting in the training phase of the models which successfully allows it to generalise to an imbalanced prediction scenario \cite{blagus2013smote}. Once the models were optimised, they were compared using their average validation set performance and finally their generalisation capability as evidenced by the test set metrics. The metrics used included Area-Under-Receiver-Operating-Curve (AUROC or AUC), Sensitivity, and Average Precision (AP) as they most completely capture the predictive performance of these binary classifiers even in cases of class imbalance. Details on how the metrics are calculated can be seen in the Supplementary Materials. \\

\noindent The XMI-ICU framework uses an extreme gradient-boosting approach with rolling time windows to extract the relevant features at defined times. This is a low-cost, time-efficient, imbalance-robust, and interpretable framework of dynamically predicting outcomes without relying on complex transfomer models for time-series analysis. The benefits of transforming a time-series dynamic prediction into n-time-window static predictions for each time point are highlighted in the Supplementary. \\

\begin{figure*}[t]
    \centering
    \includegraphics[width=0.8\linewidth]{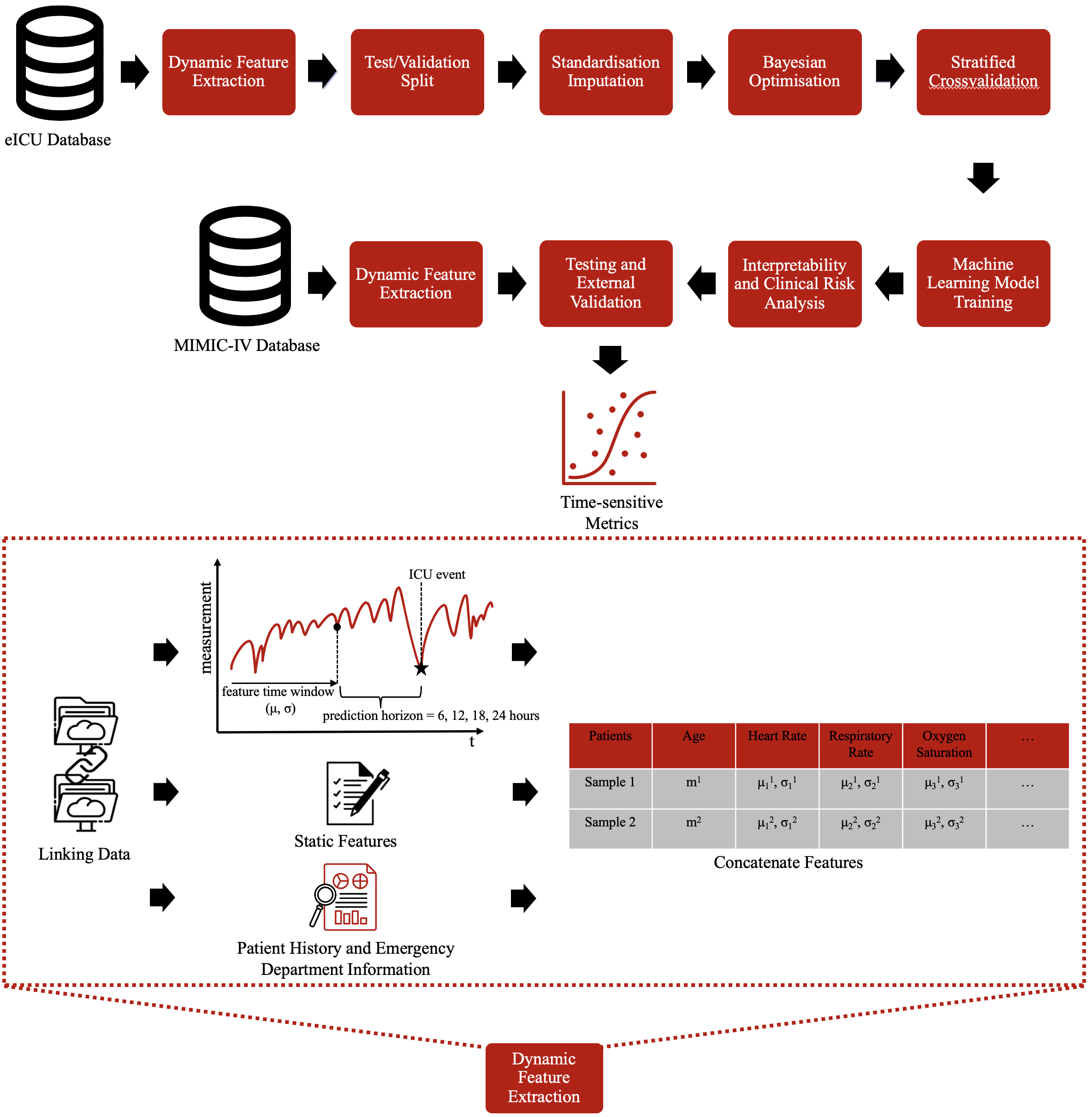}
    \caption{Proposed XMI-ICU framework For dynamic mortality prediction in heart attack ICU Patients. The top part of the figure shows the dynamic feature extraction that links hospital-wide data including pre-admission information, ICU stay measurements, and emergency department variables. The sliding time windows change depending on the required prediction time and the time-series values are summarised using mean and standard deviations. For example, for 24-hour prediction, we use all time-series measurements since time of admission until 24 hours prior to the event as our feature time window to be summarised. The measurements are then concatenated with anamnesis, emergency department, and static variables to construct the feature matrix. The bottom half of the figure showcases the framework and how the dynamic feature extraction integrates with other components.}
    \label{fig:Framework}
\end{figure*}

\noindent For time-series measurements, we leverage the advantage of gradient boosting performance previously established on tabular data. We extracted summary features like the mean and standard deviation (to preserve the units necessary for later interpretability) for each patient stay for each feature for each time window. A time window is defined as all time-series measurements for the patient during the specific stay from the time since admission to the x-time prior to the event of interest. For example, for 24-hour prediction, we use all measurements since admission into the ICU stay until 24 hours prior to the event to model a prediction scenario. Using inverse class-weighting and a sliding time window, this framework enables users or clinicians to obtain estimates for risk of death at varying times in the future through dynamic feature extraction. For each time window, features are defined with means and standard deviations. For our experiments, the time-varying performance during different time window trials of 6, 12, 18, and 24 hours prior to death was evaluated, and Shapley values were used to ascertain interpretability of the model for each of the time windows and comment on clinical significance of risk factors (also evaluated over time periods) \cite{ibrahim2020explainable}. A flowchart visualising the proposed framework for mortality prediction in MI patients can be seen in Figure \ref{fig:Framework}.\\ 

\noindent As far as the methodology of the deep learning models applied, TabNet and NODE, is concerned, more information can be found in the Supplementary Materials but a brief description is provided here. TabNet reshapes static features using sequential splits of features with a transformer and then applying attention on these sequences at each decision step to determine the most impactful features \cite{arik2021tabnet}. In effect, these sequential steps are feature selections on an instance-level repeated for each sample with a single module used for both feature selection and reasoning. A useful attribute of TabNet is also its self-supervised learning option which allows one to pre-train a TabNet model on a certain fraction of features of the data before being used for training on the full model. Initial results show that this helps increase performance on small datasets (albeit larger than ours) which we will also investigate here.\\

\noindent NODE is another alternative to the supremacy of gradient-boosted models for tabular data which relies on similar hierarchical representation learning. In short, NODE extends the idea of gradient boosting on oblivious decision trees or decision tables and it does so by making splitting features and decision tree routing differentiable. It then adds an entmax transformation which maps a vector of real values to a discrete and sparse probability distribution. This addition allows differentiable split decision construction in the internal tree nodes leading to higher efficiency and resistance to overfitting \cite{popov2019neural}. The output is a sum of leaf responses scaled by the choice weights. \\

\noindent We used the same training, validation, and held-out test sets across all of our models with the hyperparameter search space included in the Supplementary Material with the selected best-performing parameters in bold. We use standard deviation to denote the variance in our validation results across the board. Our analysis was completed in Python 3.8 using Jupyter, pandas, numpy, SHAP, the original TabNet implementation, and the extension of the NODE implementation in PyTorch Tabular with some modifications. Decision curves, clinical risk calculations, and nomograms were computed and plotted in R. \\

\subsection{Clinical risk analysis}
\noindent To provide additional analysis of the model, we used clinical impact and decision curves in estimating the performance of the model at various risk thresholds. While decision curves are mostly used in cases of intervention effect on prognosis, they can also be used to diagnose the performance of predictive models albeit their adoption in machine learning has not been widespread, possibly due to applied machine learning work in healthcare being based more on advances in computer science rather than clinical significance. Decision curves  account for both the benefits of higher risk estimation and the costs of overestimating risk to a patient who cannot benefit from the prediction. They are suggested to be an improvement over measures of performance such as AUROC. The intuition behind them is if a risk model tends to identify cases as high risk without falsely identifying too many negatives as high risk, then the net benefit of the risk model to the population will be positive \cite{kerr2016assessing}. A mathematical representation can be seen in the equation bellow:

\begin{equation}
N B_R=T P R_R P-\frac{R}{1-R} F P R_R(1-P)
\end{equation}

\noindent Where NB is the net benefit, TPR and FPR are the positive rates, and P is the prevalence and R is the risk threshold respectively. They allow us to evaluate the models across a range of risk thresholds and observing tendencies of the model to overestimate risk. A clinical impact curve is simpler in that it displays the estimated number of people declared high‐risk for each risk threshold, and visually displays the proportion of cases (true positives) \cite{chen2022development}. \\ 

\section{AUTHOR CONTRIBUTIONS}
\noindent MM, PW, and TZ designed the experiments, defined the patient criteria and outcomes. MM obtained access to, preprocessed, and cleaned the data; MM developed code for proposed model and conducted the experiments; MM, PW, and TZ contributed to the analyses of the data; MM and TZ wrote the manuscript and TZ provided overall supervision to MM. \\

\section{CONFLICT OF INTEREST STATEMENT}
\noindent None declared. \\

\section{DATA AVAILABILITY}
\noindent The data is available from public access requests for eICU and MIMIC-IV. Accessing the data requires ethics module training and certification. \\

\section{FUNDING}
\noindent M. Mesinovic appreciates the support of the EPSRC Center for Doctoral Training in Health Data Science (EP/S02428X/1) and the Rhodes Trust. \\

\ifCLASSOPTIONcaptionsoff
  \newpage
\fi

\bibliographystyle{vancouver} 
\bibliography{mortality_only}

\end{document}